\documentclass[letterpaper, 10 pt, journal, twoside]{ieeetran}
\usepackage{amsmath,amsfonts}
\usepackage{algorithmic}
\usepackage{algorithm}
\usepackage{array}
\usepackage[caption=false,font=normalsize,labelfont=sf,textfont=sf]{subfig}
\usepackage{textcomp}
\usepackage{stfloats}
\usepackage{url}
\usepackage{verbatim}
\usepackage{graphicx}
\usepackage{cite}
\usepackage{caption}
\usepackage[colorlinks=true, urlcolor=blue, linkcolor=red]{hyperref}
\newcommand{\coolname}{SceneComplete}

\newcommand\blfootnote[1]{
\begingroup
\renewcommand\thefootnote{}\footnote{#1}
\addtocounter{footnote}{-1}
\endgroup
}

\begin{document}


\title{SceneComplete: Open-World 3D Scene Completion in Cluttered Real World Environments for Robot Manipulation}

\author{Aditya Agarwal$^{1}$, Gaurav Singh$^{2}$, Bipasha Sen$^{1}$, Tomás Lozano-Pérez$^{1}$, Leslie Pack Kaelbling$^{1}$

\thanks{Manuscript received: June 14, 2025; Revised:
September 11, 2025; Accepted: October 14, 2025.}%
\thanks{This paper was recommended for publication by
Editor Júlia Borràs Sol upon evaluation of the Associate Editor and Reviewers’
comments.}%
\thanks{We gratefully acknowledge support from NSF grant 2214177; from AFOSR grant FA9550-22-1-0249; from ONR MURI grants N00014-22-1-2740 and N00014-24-1-2603; from the MIT Quest for Intelligence; and from the Robotics and Artificial Intelligence Institute.}%
\thanks{$^{1}$Aditya Agarwal, Bipasha Sen, Tomás Lozano-Pérez, and Leslie Pack Kaelbling are with Computer Science and Artificial Intelligence Laboratory, MIT. {\tt\small \{adityaag, bise, tlp, lpk\}@mit.edu}}
\thanks{$^{2}$Gaurav Singh is with the Department of Computer Science, Brown University {\tt\small gaurav@brown.edu}}
\thanks{Digital Object Identifier (DOI): see top of this page.}
}

\markboth{IEEE ROBOTICS AND AUTOMATION LETTERS. PREPRINT VERSION. ACCEPTED October, 2025}%
{Agarwal \MakeLowercase{\textit{et al.}}: SceneComplete}




\twocolumn[{
\renewcommand\twocolumn[1][]{#1}
\maketitle
\begin{center}
\vspace{-12mm}
    \centering
    \captionsetup{type=figure}
    \includegraphics[width=\textwidth]{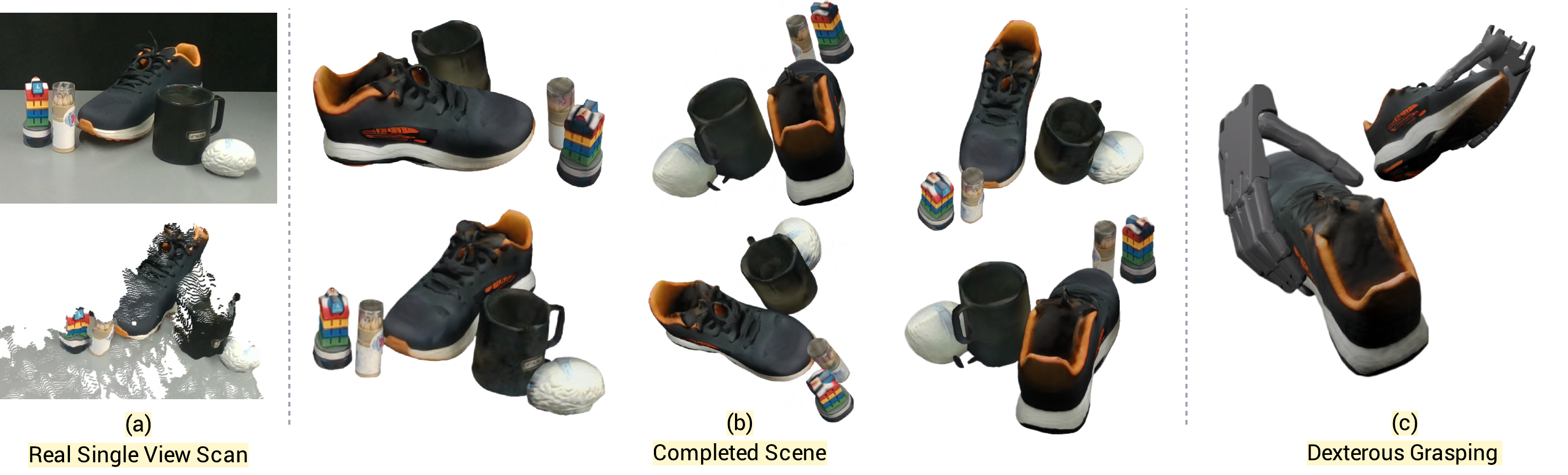}
    \captionof{figure}{(a) takes as input a {\em single} RGB-D image of a given scene, visualized here as a point cloud; (b) produces high-quality fully completed, accurately segmented object meshes in scenes with substantial occlusion and novel objects; and (c) enables downstream dexterous manipulation that requires accurate complete shape information.}
\end{center}
}]


\blfootnote{
{Manuscript received: June 14, 2025; Revised:
September 11, 2025; Accepted: October 14, 2025}

{This paper was recommended for publication by
Editor Júlia Borràs Sol upon evaluation of the Associate Editor and Reviewers’
comments.}

{We gratefully acknowledge support from NSF grant 2214177; from AFOSR grant FA9550-22-1-0249; from ONR MURI grants N00014-22-1-2740 and N00014-24-1-2603; from the MIT Quest for Intelligence; and from the Robotics and Artificial Intelligence Institute.}

{$^{1}$Aditya Agarwal, Bipasha Sen, Tomás Lozano-Pérez, and Leslie Pack Kaelbling are with Computer Science and Artificial Intelligence Laboratory, MIT. {\tt\small \{adityaag, bise, tlp, lpk\}@mit.edu}}

{$^{2}$Gaurav Singh is with the Department of Computer Science, Brown University. {\tt\small gaurav@brown.edu}}

{Digital Object Identifier (DOI): see top of this page.}
}

\vspace{-10px}
\begin{abstract}
Careful robot manipulation in every-day cluttered environments requires an accurate understanding of the 3D scene, in order to grasp and place objects stably and reliably and to avoid colliding with other objects. In general, we must construct such a 3D interpretation of a complex scene based on limited input, such as a single RGB-D image.  We describe \coolname{}, a system for constructing a complete, segmented, 3D model of a scene from a single view. \coolname{} is a novel pipeline for composing general-purpose pretrained perception modules (vision-language, segmentation, image-inpainting, image-to-3D, visual-descriptors and pose-estimation) to obtain highly accurate results. We demonstrate its accuracy and effectiveness with respect to ground-truth models in a large benchmark dataset and show that its accurate whole-object reconstruction enables robust grasp proposal generation, including for a dexterous hand. We release the code and additional results on our \href{https://scenecomplete.github.io}{website}.
\end{abstract}


\vspace{-5px}
\section{Introduction}

\IEEEPARstart{A}{s} manipulation robots move from constrained environments such as factories and workshops to open-world environments such as homes and hospitals, they must be able to construct representations of their environment that enable robust, careful manipulation. Such representations need to individuate objects and characterize their shapes, so that the robot can reliably select stable grasps and placements for individual objects and manipulate them without unwanted collisions.  These representations must generally be constructed from limited input, such as a single RGB-D image.
This problem is fundamentally ill-posed, but we are now in a position to address it using strong priors that have been learned by vision foundation models.

In this paper, we propose a solution to this open-world scene completion problem in the form of a perception pipeline, \coolname{}. It combines multiple large pre-trained vision models into a system that takes a single RGB-D image as input and predicts as output a completed scene, consisting of a set of meshes for all the visible objects, including those that are partially occluded. Crucially, it makes no assumptions about the categories of the objects, their arrangement, or the camera viewpoint.  It is constructed from multiple highly capable pre-trained perception components: a vision-language model (VLM) for identifying and generating short descriptions of the objects in a scene, a text-grounded image-segmentation model for localizing objects in the image, a 2D image-inpainting model for predicting the appearance of occluded parts of objects, an image-to-3D model for generating complete object meshes, and visual descriptor and pose-estimation modules to aid in composing individual predicted meshes into a final scene.
None of these components can individually solve the problem, but in combination they provide robust object-centric interpretation of complex images, producing a segmented set of object meshes that are suitable for robot planning and manipulation.

To demonstrate the effectiveness of \coolname{}, we conduct extensive  quantitative and qualitative evaluations on real-world tabletop scenes. Our quantitative evaluations are on the Graspnet-1B~\cite{graspnet-1b} and YCB-Video~\cite{xiang2017ycbv} datasets, which consist of cluttered tabletop scenes. In these scenarios, accurately predicting the full scene—--including partially occluded objects—--is crucial for stable grasping, collision-free motion with an object in the hand, and reliable placing. 

We further illustrate the utility of our shape-reconstruction methods by using them as input to parallel-jaw~\cite{dexnet, contact-graspnet} and dexterous grasping~\cite{dexgraspnet} methods, the latter being especially sensitive to the detailed shape {\em of the entire object}.

\begin{figure*}[t]
    \centering
    \includegraphics[width=1\textwidth]{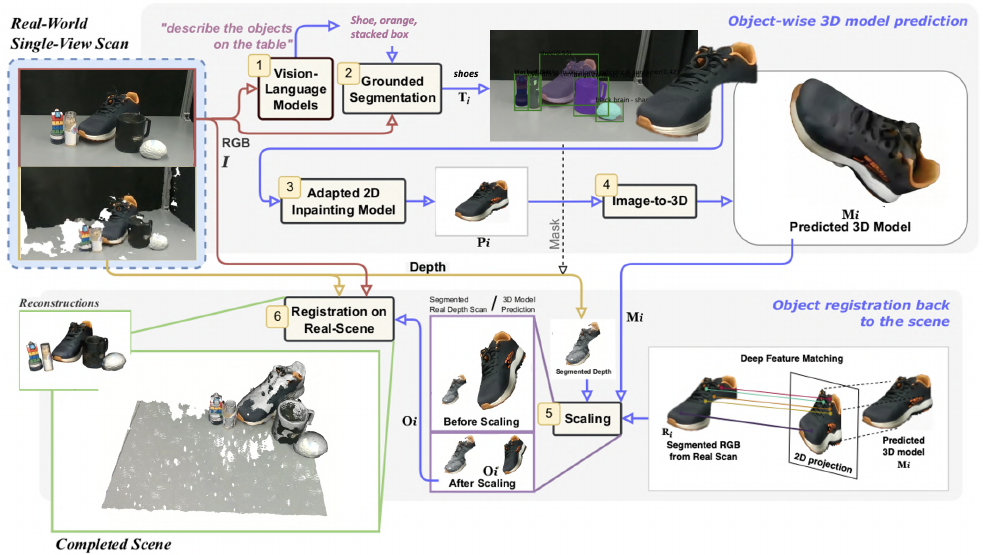}
    \caption{\textbf{Overview of the SceneComplete pipeline. }Starting from a single RGB-D input, the system produces a set of object meshes registered with the input 3D scan, yielding a complete 3D scene reconstruction. The pipeline consists of six key phases: (1) An RGB image is fed into a VLM to enumerate and describe objects, (2) object descriptions and the RGB image are processed by a grounded segmentation model to generate object masks, (3) occluded regions are completed via image inpainting model adapted to output single fully observable objects on a white background, (4) the inpainted 2D images are passed into an image-to-3D model to produce object meshes, (5) object meshes are scaled according to the segmented partial point cloud, and (6) mesh poses are adjusted within the 3D coordinate frame of the original scan using 6DOF pose estimation. Each step leverages pre-trained open world large vision models, enabling scalability and benefiting from future model improvements.}
    \label{fig:pipeline}
    \vspace{-14px}
\end{figure*}

\vspace{-10px}
\section{Related Work}
\textbf{Feed-forward Scene Reconstruction: }Feed-forward multi-object scene completion methods such as ~\cite{lunayach2024fsd, irshad2022shapo, iwase2024octmae, iwase2025zerograsp, shi2025crisp} learn end-to-end mappings from single-view RGB-D input to completed object meshes or occupancy grids. These methods are attractive because they are computationally fast at inference time; however, suffer from key limitations: ShAPO~\cite{irshad2022shapo}, FSD~\cite{lunayach2024fsd}, and CRISP~\cite{shi2025crisp} are closed-set methods, trained extensively on fixed benchmark datasets with predefined object categories, and fail to generalize to novel objects. OctMAE~\cite{iwase2024octmae} addresses open-set scenes by performing scene-level reconstruction. However, its predictions are surface-level and do not individuate objects, limiting their utility for robotic manipulation. ZeroGrasp~\cite{iwase2025zerograsp} performs scene reconstruction with individuated objects for grasp generation, but the reconstruction quality—crucial for collision avoidance—is subpar since the focus is primarily on grasping. In our results, we compare and outperform OctMAE and ZeroGrasp in reconstructing scenes and generating grasps.

\textbf{Compositional Scene Reconstruction: }This line of work composes multiple open-set models with zero-shot generalization into a pipeline for scene reconstruction. While such methods are typically slower than feedforward methods, they can directly handle unseen objects with little or no retraining~\cite{yao2025cast, ardelean2024gen3dsr, 10802733}. CAST~\cite{yao2025cast} and Gen3DSR~\cite{ardelean2024gen3dsr} both reconstruct full scenes from a single RGB image by first predicting depth maps from monocular depth estimation models. CAST then generates meshes for individual objects and applies a physics-aware correction step to enforce physically consistent placements, while Gen3DSR integrates components such as DreamGaussian to produce full 3D meshes. However, because both methods rely on predicted depth, their pipelines are tightly coupled to synthetic outputs and fail to adapt to raw RGB-D input, where sensor noise is unavoidable. These approaches are primarily designed for asset generation, and their inability to incorporate ground-truth depth limits their applicability in robotics. By contrast, \coolname{} leverages observed sensor depth (often noisy), enabling a more flexible pipeline tailored for manipulation tasks. 

A different line of work, Open6DOR~\cite{10802733}, introduces a benchmark for language-driven 6-DoF object rearrangement, along with a baseline pipeline for grasp generation. Their approach composes a set of modules similar to those in SceneComplete, but focuses exclusively on predicting object poses. As a result, the pipeline does not perform full scene reconstruction, omits critical steps such as registration, and restricts predictions to objects that are almost fully visible.


\vspace{-10px}
\section{Method}
\label{sec:approach}

\setlength{\abovedisplayskip}{0pt}
\setlength{\belowdisplayskip}{3pt}



\newcommand{\obs}{I}           
\newcommand{\orgb}{\obs^{\text{rgb}}}  
\newcommand{\odep}{\obs^{\text{depth}}} 
\newcommand{\opos}{\theta}  
\newcommand{\omask}{\mathbf{m}} 
\newcommand{\mfeat}{\mathbf{f}} 
\newcommand{\objf}{\mathbf{f_o}} 
\newcommand{\mbsp}{\mathbf{p}}  
\newcommand{\objp}{\mathbf{p_o}} 

\newcommand{\ibox}{\mathbf{b}} 
\newcommand{\pbox}{\ibox^{\text{3D}}} 

\newcommand*{\seg}{\ensuremath{\text{Seg}}} 
\newcommand{\SE}{\text{Embed}}        
\newcommand{\lang}{\text{LVLM}}        
\newcommand{\glang}{\text{LLM}}        

\newcommand{\objmap}{\mathcal{M}} 
\newcommand{\obj}{\mathbf{o}}      
\newcommand{\objset}{\mathbf{O}}   
\newcommand{\edg}{\mathbf{e}}      
\newcommand{\edgeset}{\mathbf{E}}  
\newcommand{\capp}{\mathbf{c}}      
\newcommand{\rcapp}{\hat{\capp}}      

\newcommand{\semSim}{\phi_{\text{sem}}}  
\newcommand{\geoSim}{\phi_{\text{geo}}}  
\newcommand{\nnratio}{\text{nnratio}}  
\newcommand{\overallSim}{\phi}          
\newcommand{\simMat}{\mathbf{\phi}}     
\newcommand{\newofeat}{\objf_{\text{new}}}  
\newcommand{\bbodt}{\epsilon_b} 
\newcommand{\bbmat}{\mathbf{\psi}} 
\newcommand{\oldofeat}{\objf_{\text{old}}}  
\newcommand{\numDet}{n}              
\newcommand{\sminus}{\text{-}}  

\newcommand{\imgseq}{\mathcal{I}} 

\newcommand{\wedgev}{w}      
\newcommand{\ovThresh}{\alpha}  

\newcommand{\mstree}{\mathcal{T}}  


Figure~\ref{fig:pipeline} illustrates the overall design of \coolname{}.  It takes a single RGB-D image as input and produces a set of object meshes that are registered with the input 3D scan.  The objective is to provide an accurate 3D reconstruction of the scene, in terms of segmentation into rigid components, and the shape of each component, expressed as a mesh.  
Importantly, each step in the pipeline makes use of existing pre-trained open-world visual-processing models, with almost no additional training (we do a small amount of low-rank adaptation of the inpainting model). 
This means that, as improved models become available for each of these tasks, as they inevitably will, we will be able to immediately profit from these improvements.  
We describe each process in detail in the following subsections. 

\vspace{-11px}
\subsection{Prompting and segmentation}
\label{sec:promptandseg}

We begin by using a vision-language model to determine the number and basic description of objects in the scene. In our implementation, we pass the RGB image $I$ into ChatGPT-4o\footnote{https://openai.com/index/hello-gpt-4o/} with the prompt \textit{``describe the objects in the image with their generic name and color as prompts in a list."} It produces a text response as a list $t_1, \ldots, t_n$ of text descriptions of objects. E.g., in the example shown in Fig.~\ref{fig:pipeline}, it returns \textit{``Blue Bowl, Tape, Banana"} and so on. 

Next, we obtain an image mask for each object. For each text description $t_i$, we prompt a grounded segmentation model (in our implementation, GroundedSam2~\cite{groundedsam}) using $t_i$ on image $I$ and obtain a candidate set of pairs of masks and confidence values.   
For example, the prompt ``a pear'' might return multiple useful masks in a scene with two pears; in other cases, some of the masks will be unhelpful (and hopefully low-confidence).  We use the confidence values to greedily select a set of non-overlapping masks, and associate each mask with the text prompt that generated it, producing the set $(R_1, T_1), \ldots, (R_N, T_N)$, $R$ and $T$ denoting the masks and prompts, respectively, for $N$ objects in the scene.  

\vspace{-11px}
\subsection{Image Inpainting}
\label{sec:inpainting}

\begin{figure}[H]
    \centering    
    \vspace{-10px}
    \includegraphics[width=0.95\linewidth]{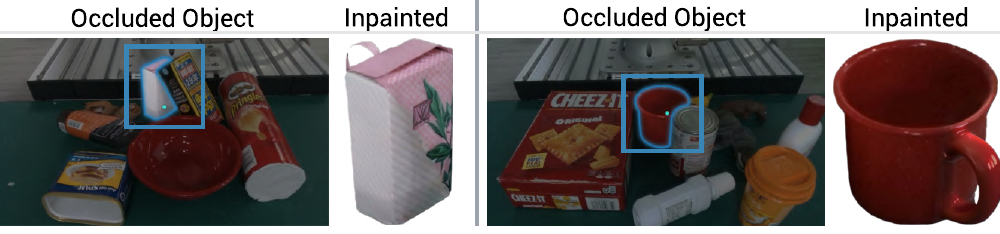}
    \caption{In the image inpainting module, occluded objects (blue borders) are transformed into single fully observable objects.}
    \vspace{-10px}
    \label{fig:inpaintingsmallex}
\end{figure}
\begin{figure*}[t]
    \centering
    \includegraphics[width=0.95\linewidth]{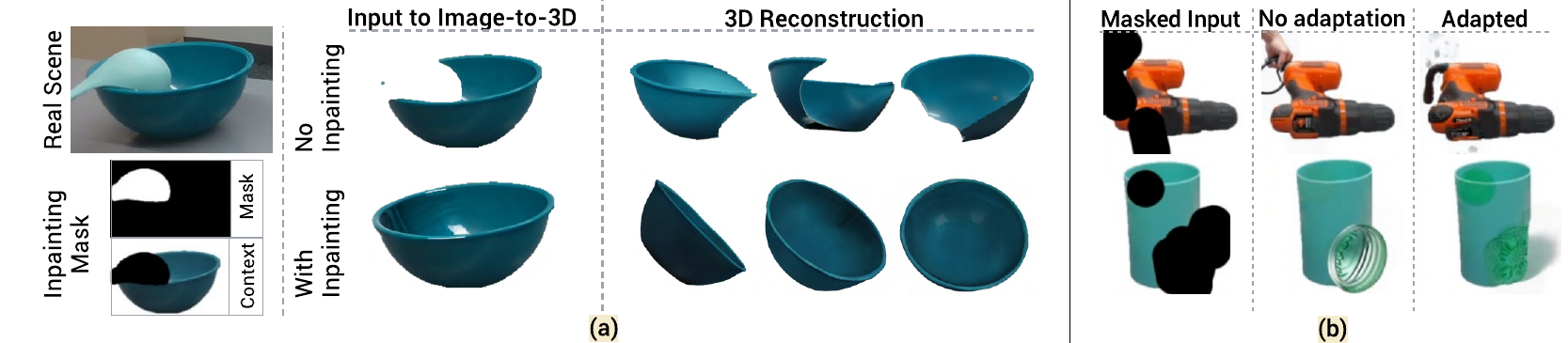}
    \caption{(a) The impact of inpainting on image-to-3D reconstruction. Without inpainting (top), the image-to-3D model generates incomplete meshes. Inpainting (bottom) fills in occluded parts, producing accurate 3D reconstructions. (b) Comparison of inpainting models. Unadapted BrushNet (middle) introduces artifacts, while the adapted version (right) inpaints occluded parts correctly producing a fully observed object.}
    \vspace{-20px}
    \label{fig:inpaintingvsnoinpainting}
\end{figure*}

As illustrated in the Fig.~\ref{fig:inpaintingsmallex}, the objects represented by the masks in $(R_1, T_1), \ldots, (R_N, T_N)$ could be partially occluded by other objects in the scene. In this step, we use an image inpainting algorithm to fill in the occluded parts of each object's image. This is important for the next step of predicting a 3D mesh, because the image-to-3D models only performs reliably with a complete view of the object. Fig.~\ref{fig:inpaintingvsnoinpainting} (a) illustrates the poor results of attempting reconstruction from an incomplete image.

In our implementation we begin with BrushNet~\cite{brushnet}, which takes an image with explicitly masked out regions and a text prompt, and produces a completed image, with the masked portions filled in. So for each $(R_i, T_i)$ pair, we begin by constructing an image $I_i$ with just the segmented object region $R_i$ on a white background. Then we construct an inpainting mask consisting of the union of the regions $R_j$ for $j \not = i$.  Finally we query Brushnet with $I_i$, $\cup_{j \not= i} R_j$, and $T_i$
and obtain an inpainted image $P_i$ of the completed object.

However, we observed that, out of the box, BrushNet occasionally synthesized additional objects in the occluded areas, as shown in Fig.~\ref{fig:inpaintingvsnoinpainting} (b), possibly due to the ``artistic'' data set on which it was trained.
To improve this behavior, we adapt BrushNet to full, single objects. It is important to note that, even though we want to adapt BrushNet to \textbf{single} \textbf{fully observed} objects, we want to retain the open world capabilities of the model. To achieve this, we use Low-Rank Adaptation (LoRA)~\cite{lora} on its learnable layers with the PEFT method, targeting domain-specific improvements on the tabletop YCB dataset~\cite{ycbdataset}. LoRA is an effective method for adapting pretrained models to domain-specific outputs while retaining the inherent generalization of the models. 
To perform this adaptation, we project the different 3D YCB object meshes from arbitrary poses on a white background and add random brush masks for inpainting as suggested by BrushNet. 
The adapted model allows us to reliably inpaint individual occluded objects, including those from categories not in the adaptation dataset. 

\vspace{-10px}
\subsection{Image-to-3D models for object reconstruction}
\label{sec:im23D}
At this point, we have, for each object, a fully observed object image $P_i$ on a white background. 
The next step is to generate a 3D mesh model for the object. Although methods exist for operating directly on the point-cloud generated from the depth channel of the RGB-D image, the depth information is often very low-resolution and noisy, so these depth-only models tend to be highly tuned to specific categories~\cite{gan_inversion, scarp, irshad2022shapo, convoccupancynetworks} and viewpoints, or operate over idealized sensory input. In recent years, substantial improvements have been made in RGB-only methods, which take advantage of the high quality of the RGB signal and the enormous amounts of available training data~\cite{zero123, one-2345, instantmesh}.  

For RGB image as inputs, one option is to use methods that optimize a 3D mesh through differentiable rendering~\cite{dreamfusion, score-jacobian-chaining, 3dfuse}.
Although these models have impressive open world results, they have substantial computational cost. On the other hand, feed-forward methods such as InstantMesh~\cite{instantmesh} that directly map a single-view RGB image into a 3D mesh are highly accurate, open world, and computationally inexpensive at the time of inference.
For these reasons, we use InstantMesh, providing each image $P_i$ as input and obtaining a complete textured 3D mesh $M_i$ as output. The resulting meshes are produced in an arbitrary orientation, at an arbitrary scale, so more work remains to be done, as explained in the subsequent subsections.

\vspace{-12px}
\subsection{Mesh Scaling using Dense Correspondence Matching}
\label{sec:scaling}
The next step is to rescale the meshes $M_i$, using a point-cloud constructed from the region $R_i$ to determine the scale factor as follows:
(i) Using the default viewpoint $v$ from InstantMesh, generate an image $V_i$ as a 2D projection of mesh $M_i$.  
(ii) Following~\cite{deepdescriptors}, find dense visual descriptors of the segmented original image, $R_i$ and the projected image, $V_i$, using a pre-trained vision transformer. In our observation, $v$ is usually close to the viewpoint from which $R_i$ was rendered. This enables us to obtain matching visual descriptors across $R_i$ and $V_i$.  
(iii) Generate dense pixel-wise correspondences between the descriptor images, so that we have a set of pairs $(R_i^j, V_i^j)$ where $R_i^j$ is a pixel in the original image of the object and $V_i^j$ is a pixel in the synthesized view.  
(iv) Map these correspondences into 3D, to obtain a pair $(R_i^{j'}, V_i^{j'})$ of points in 3D. (Note that these point sets may not yet be aligned in 3D---we address that problem in the next step.)  
(v) Center each resulting 3D cloud, compute the average Euclidean distance from the points to the centroid, and compute the ratio of these values as the scale factor.  
(vi) Apply the scale factor to the mesh $M_i$ to put it in the same scale as the original point cloud to obtain $O_i$ as shown in Fig.~\ref{fig:pipeline}-(5).

\vspace{-12px}
\subsection{6D Object Pose Estimation and Registration}
\label{sec:poseestandregis}
Now, we have, for each object, an appropriately scaled mesh $O_i$, and we need to reconstruct the entire scene. To do this, we need to find a 6DOF transform for each mesh that causes it to register well with the observed point cloud.
For this task, we use FoundationPose~\cite{foundationpose}, a robust object-pose estimation method designed to operate without being limited to specific object categories. We use its model-based mode, which takes as input a partial point-cloud derived from region $R_i$ of the input RGB-D image and an appropriately scaled textured object mesh $O_i$, and returns a 6DOF transform $\tau_i$ mapping $O_i$ into the coordinate frame of the point cloud.

As a result of this process, we have a set of pairs $(O_i, \tau_i)$, where each $O_i$ is a textured complete mesh for object $i$ and $\tau_i$ is a pose for that mesh in the camera coordinate frame.  This scene reconstruction provides a highly general representation for a wide variety of downstream object-manipulation tasks. Accurate reconstructions of unobserved parts of objects enables a wide variety of manipulation operations, including  many types of robust grasping, moving safely in cluttered but unobserved parts of the scene, moving safely when holding a grasped object, etc.

\vspace{-10px}
\section{Experiments}
\label{sec:results}

\begin{figure*}
    \centering
    \includegraphics[width=0.88\textwidth]{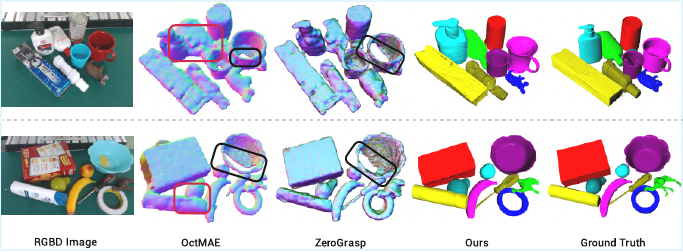}
    \vspace{-4px}
    \caption{\textbf{Qualitative comparisons} of scene reconstructions on the GraspNet-1B dataset. For each scene we show, the input RGB-D image, OctMAE reconstruction (rendered as normal maps as it predicts scene-level occupancy values), ZeroGrasp reconstruction (rendered as normal maps), our reconstruction (visualized as individually reconstructed object meshes color-matched to the ground truth), and ground-truth object meshes. Highlighted regions indicate missing area (black) or spurious region connecting distinct objects (red).}
    \label{fig:gs1b_reconstruction_comparisons}
    \vspace{-10px}
\end{figure*}

\begin{figure}[t]
    \centering
    \noindent\includegraphics[width=\linewidth]{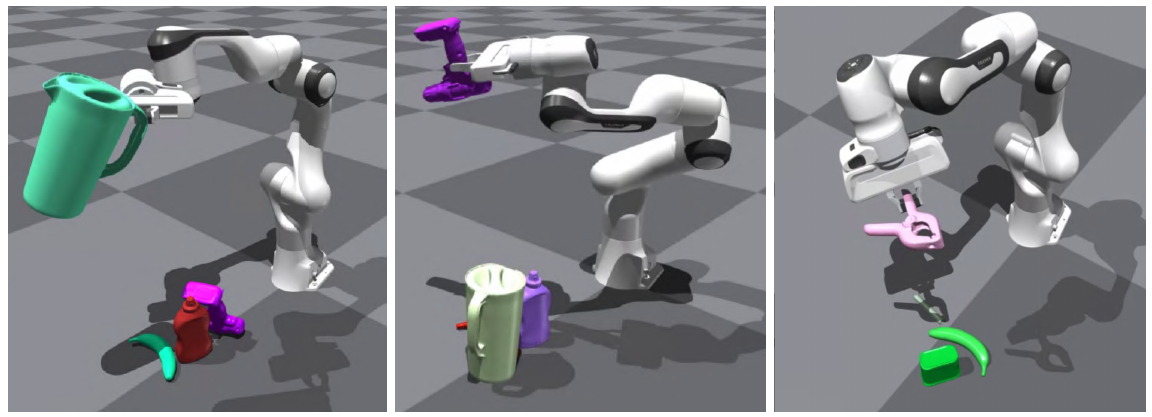}
    \caption{Evaluating \textit{GSR} on the YCB-V dataset in Isaac Gym. We show a close-up of 3 distinct objects being picked up. }
    \label{fig:ycbv_evaluation}
\end{figure}

\begin{figure}[t]
    \centering
    \noindent\includegraphics[width=\linewidth]{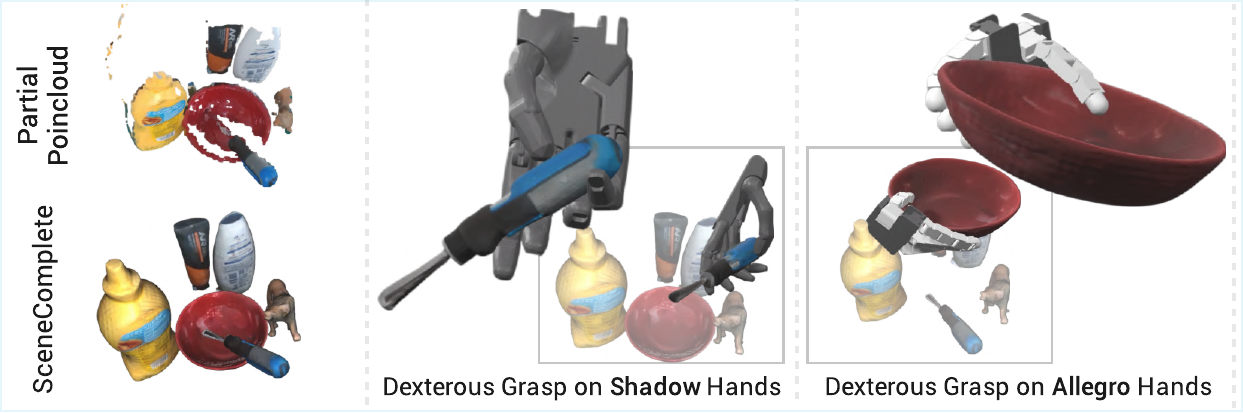}
    \caption{We demonstrate dexterous grasps using both Shadow Hands and Allegro Hands~\cite{dexgraspnet} on objects from the GraspNet-1B dataset, highlighting improved manipulation of complex objects with complete 3D reconstructions.}
    \label{fig:dexterous_grasping}
    \vspace{-18px}
\end{figure}

\vspace{-2px}
We evaluate \coolname{} through three main experimental regimes: 
\begin{itemize}
    \item \textbf{Scene Reconstruction and Grasping}: We compare \coolname{} against an existing single-view scene reconstruction method on a large-scale dataset of tabletop scenes, GraspNet-1B~\cite{graspnet-1b}, captured using a RealSense D435 camera. We also evaluate how the reconstructed scenes contribute in generating collision-free grasps. 
    \item \textbf{Object-grasping and Dexterous Manipulation}: We evaluate the effectiveness of \coolname{} in generating grasps for successfully picking up objects inside a simulation environment using a parallel-jaw gripper. We also demonstrate that the reconstructed object models are of sufficient fidelity to enable dexterous grasp proposals for a multi-fingered hand, which depends on having a good estimate of the entire object shape. 
    \item \textbf{Real-world Evaluations}: To assess the real-world applicability of \coolname{}, we conduct pick-and-place experiments on a physical robot on a smaller-scale dataset of tabletop scenes collected in our lab (also using a RealSense D435 camera), that include everyday objects which are less likely to have appeared in the training distribution of any of the models used. We evaluate across 15 scenes, each with 4 to 6 objects. 
\end{itemize}

We observe that the runtime of the current implementation of \coolname{} is about 20s per object on a single NVIDIA RTX 4090 GPU; we expect this to improve with advances in models and GPU architectures. 




\vspace{-10px}
\subsection{Scene Reconstruction and Grasping}
We first evaluate \coolname{} on the task of reconstructing tabletop scenes from just a single RGB-D image. 

\textbf{Dataset}: We use the GraspNet-1B dataset~\cite{graspnet-1b} for our evaluation. This dataset is particularly suitable for our setting, due to its large collection of 190 cluttered tabletop scenes, featuring 88 unique objects in various configurations. It includes ground-truth 3D object models and poses for each scene, as well as real RGB-D images.



\textbf{Baseline}: Our primary baselines for comparison are OctMAE~\cite{iwase2024octmae} and ZeroGrasp~\cite{iwase2025zerograsp}. OctMAE performs reconstruction by combining octree-based representation with a 3D Masked AutoEncoder (MAE). ZeroGrasp performs simultaneous 3D reconstruction and 6D grasp pose prediction using an octree-based CVAE. Both methods expect a single-view RGB-D image along with a corresponding foreground mask as input, and output the reconstructed scene. 
For our experiments, we provide OctMAE and ZeroGrasp with the ground-truth segmentation masks.
We also report numbers against the partial point cloud (PartialDecomp).

\textbf{Metrics}: To measure 3D reconstruction quality and fidelity, we report well-known 3D metrics such as \textit{Chamfer distance (CD)} and \textit{Earth Mover's distance-maximum mean discrepancy (MMD-EMD)} metric, similar to existing works~\cite{iwase2024octmae, scarp, shu20193dtreegan}.
Both CD and MMD-EMD metrics expect pointclouds as input. For \coolname{}, we sample points uniformly from the reconstructed object meshes. Since OctMAE predicts occupancy values, we use the reconstructed point cloud produced by its occupancy values, normal vectors, and SDF. 
We also report the \textit{Mesh Intersection-over-Union (MIoU)} metric, which is based on comparing the ground-truth meshes with the reconstructed meshes. Specifically, let $U^*$ be the union of volumes enclosed by the ground truth object meshes, and let $\widehat{U}$ be the union of the volumes enclosed by the meshes produced by a reconstruction algorithm. Then the intersection-over-union metric between the meshes is
\[\text{MIoU}(U^*, \widehat{U}) = \frac{U^*\cap \widehat{U}}{U^*\cup \widehat{U}}\;\;.
\]
MIoU explicitly penalizes both under-reconstructions (missing parts), over-reconstructions (excess geometry), and registration errors. To compute MIoU, we make the reconstructed meshes produced by \coolname{}, ZeroGrasp, and OctMAE watertight using ManifoldPlus~\cite{huang2020manifoldplus}.

\textit{Collision-free Grasping}: We perform a basic test of the utility of \coolname{} on an important downstream task of grasping, by using an antipodal grasp generation method~\cite{dexnet} to generate grasps on the objects in the reconstructed scene. 
To illustrate the importance of whole-scene reconstruction on grasping, we ask the question: of these potential grasps, which ones are {\em in collision with the ground-truth scene}?  Concretely, for a reconstructed scene, we (a) sample a set $G$ of collision-free grasps using antipodal sampling, (b) evaluate grasps in $G$ that would cause a collision in the ground truth scene to obtain a subset $G'$, and compute the \textit{Grasp Collision metric GC} as ($\lvert G' \rvert / \lvert G \rvert$)
which is the percentage of \textit{grasps that the reconstruction would allow, that in fact collide}, similar to the metric adopted by~\cite{carvalho2024graspdiffusion,scarp}. In our case, $G$ is set to 40. 

\begin{table}[t]
\centering
\small
\setlength{\tabcolsep}{5pt}
\renewcommand{\arraystretch}{1.2}
\begin{tabular}{|l|ccc|c|}
\hline
& \multicolumn{3}{c|}{\textbf{Reconstruction}} & \multicolumn{1}{c|}{\textbf{Grasping}} \\
& \textbf{MIoU}$\uparrow$ 
& \textbf{CD}$\downarrow$ 
& \textbf{MMD-EMD}$\downarrow$
& \textbf{GC}$\downarrow$ \\ \hline
PartialDecomp    
& 0.166 
& 3.16
& 3.32
& 53.5 \\

OctMAE~\cite{iwase2024octmae}
& 0.445
& 1.73 
& 3.11 
& 20.3 \\

ZeroGrasp~\cite{iwase2025zerograsp}
& 0.440
& 1.86
& 3.07
& 18.9 \\

\coolname{}    
& \textbf{0.478} 
& \textbf{1.54} 
& \textbf{3.06} 
& \textbf{16.4} \\ \hline
\end{tabular}
\caption{Comparison of shape reconstruction methods on GraspNet-1B. Higher MIoU indicates better shape fidelity. Lower CD, MMD-EMD, and GC indicate more accurate and feasible reconstructions for downstream grasping. CD and MMD-EMD are scaled by $10^4$ and $10^2$ respectively.}
\label{tab:combined_metrics}
\vspace{-2em}
\end{table}

\begingroup
\begin{table*}[t]
\centering
\small
\setlength{\tabcolsep}{8pt}
\renewcommand{\arraystretch}{1.2}
\begin{tabular}{lccc}
\hline
  & \textbf{Contact-GraspNet GSR} 
  & \textbf{Antipodal Grasping GSR}
  & \textbf{Overall GSR\(\uparrow\)}\\ \hline
PartialDecomp  
  & $0.46 \pm 0.34$ 
  & $0.17 \pm 0.13$
  & 0.32 \\
SceneComplete  
  & \textbf{$\textbf{0.81} \pm \textbf{0.2}$}  
  & \textbf{$\textbf{0.73} \pm \textbf{0.18}$} 
  & \textbf{0.77} \\ \hline
\end{tabular}
\caption{Overall Grasp Success Rate (GSR\(\uparrow\)) on the YCB-V dataset for two grasping methods.}
\label{tab:pick-success}
\vspace{-2em}
\end{table*}
\endgroup

\textbf{Results}: As shown in Table~\ref{tab:combined_metrics}, our method outperforms the baseline methods in all metrics. We also visualize the comparisons in Figure~\ref{fig:gs1b_reconstruction_comparisons}. While OctMAE produces visually plausible reconstructions, it sometimes fails to recover parts of objects that are occluded by other objects, or are partially observable due to the viewpoint. This also results in higher grasp collisions as shown in Table~\ref{tab:combined_metrics}, while our method recovers such missing regions. 
Moreover, since OctMAE directly predicts a scene-level reconstruction, it often hallucinates geometry connecting distinct objects (highlighted in Fig.~\ref{fig:gs1b_reconstruction_comparisons}), which results in a lower MIoU. In contrast, our object-centric approach reconstructs each object individually, preserving a clear separation between them. 
We find that on average, only \textbf{16\%} of grasps generated by \coolname{} are in collision with real objects, as opposed to 20\% of those generated by OctMAE, 19\% of those generated by ZeroGrasp, and 53\% of those generated by PartialDecomp. 

\begin{figure*}
    \centering
    \includegraphics[width=0.95\textwidth]{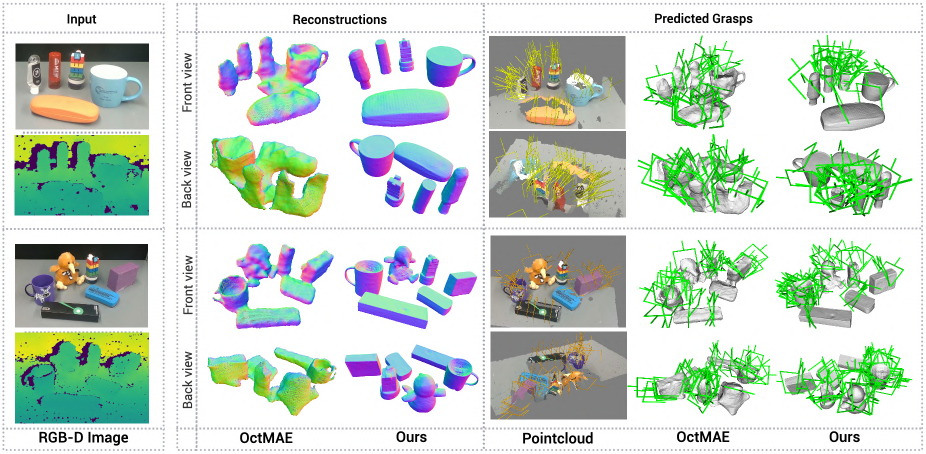}
    \vspace{-3px}
    \caption{\textbf{Qualitative comparisons} of scene reconstructions on the scans collected in our lab. For each scene we show, the input RGB-D image, OctMAE reconstructions, our reconstructions, grasp proposals on input partial point cloud, grasp proposals on OctMAE's reconstructions, and grasp proposals on our reconstructions. We show the scene from a front and back viewpoint.}
    \label{fig:lab_reconstruction_comparisons}
    \vspace{-10px}
\end{figure*}


\vspace{-10px}
\subsection{Object Grasping and Dexterous Manipulation}
We next evaluate the utility of \coolname{} for object grasping and dexterous manipulation in simulation.


\textbf{Dataset}: We use a subset of the YCB-Video~\cite{xiang2017ycbv} dataset consisting of 30 cluttered tabletop scenes with 4 to 5 objects per scene from the YCB~\cite{ycbdataset} dataset for simulated grasping experiments using a parallel jaw gripper. For dexterous manipulation, we evaluate on 20 object instances chosen arbitrarily from the GraspNet-1B~\cite{graspnet-1b} dataset. 

\textbf{Baseline: }We compare against the raw input partial point cloud (PartialDecomp) as the baseline. Since this experiment is object-centric, its not compatible with OctMAE, which produces scene-level reconstruction without per-object separation. 
While instance segmentation could in principle be used to individuate objects, they do not reliably recover the extent or shape of each reconstructed object, making fair comparison infeasible for object-centric manipulation tasks. We show comparisons with OctMAE on a scene-level grasping task in the next section. 


\textbf{Metrics}: To assess the effectiveness of \coolname{} for object manipulation, we evaluate the \textit{Grasp Success Rate (GSR)} using simulated grasp attempts (a common metric adopted by prior works~\cite{zhang2024graspxl, jiang2021synergies_giga}) in Isaac Gym. Our scene consists of a Franka Emika Panda arm with a parallel jaw gripper, and scenes from the YCB-V dataset~\cite{xiang2017ycbv}. We compare the GSR achieved by \coolname{} against PartialDecomp and adopt two established and distinct grasping methods for generating grasp proposals: 
\begin{itemize}
    \item Antipodal grasping using~\cite{dexnet}: This method samples antipodal points to generate collision-free grasps on the object meshes. To make the objects compatible with simulation and to improve efficiency, we perform an approximate convex decomposition of the watertight object meshes into convex hulls using CoACD~\cite{wei2022coacd}.
    \item Contact-GraspNet~\cite{contact-graspnet}: This method generates grasp proposals directly from point clouds and was trained to predict grasps from partial observations. We generate grasp proposals on both the reconstructed point cloud (from \coolname{}) and the raw partial point cloud, and evaluate these grasps on the ground-truth scenes. 
\end{itemize}

For each grasping method, we first generate candidate grasps on both \coolname{}'s reconstructed scene and PartialDecomp. For each grasp, we simulate a pick attempt on the reconstructed meshes in Isaac Gym and retain only those grasps for which the object remains securely held in the gripper. From these filtered grasps, we randomly select upto $40$ grasps per object and evaluate them on the ground-truth object meshes in Isaac Gym. Each evaluation consists of successfully picking up the object from its computed grasp and holding it in the air without dropping, as shown in Fig.~\ref{fig:ycbv_evaluation}. GSR is computed as the proportion of successful grasp attempts out of the $40$ evaluated grasps. Objects that do not yield valid grasps due to constraints such as exceeding the gripper width are excluded from our evaluation. 

\textit{Dexterous Grasping and Stability}: A significant test of the utility of our approach is whether the object reconstructions support the computation of good dexterous grasps for a multi-fingered hand. 
We evaluate object reconstruction as: 
\begin{itemize}
    \item Pass a {\em reconstructed} object mesh $O_i$ into DexGraspNet~\cite{dexgraspnet}, configured for a dexterous hand (we used both  Shadow and Allegro hands), to obtain a grasp $g_i$.
    \item Instantiate Isaac Gym with the selected hand and the {\em ground truth} object mesh.
    \item Similar to other methods that evaluate dexterous grasping~\cite{dexgraspnet, shao2024bimanualdexgrasp}, we lift the hand and rotate it within the simulation, and detect whether the object is dropped using PhysX as the physics engine. We visualize dexterous grasps on representative objects using both Shadow and Allegro hands in Fig.~\ref{fig:dexterous_grasping}. 
\end{itemize}
We calculate the percentage of such tests that succeed. For evaluation, we selected 20 objects from the GraspNet-1B dataset and evaluated \coolname{} against PartialDecomp on ground-truth objects—an important upper bound illustrating the reliability of DexGraspNet for selecting such grasps.


\textbf{Results}: As shown in Table~\ref{tab:pick-success}, reconstructing scenes with \coolname{} significantly improves grasp success rates in simulation using a parallel jaw gripper. Across both grasping methods, \coolname{} achieves over \textbf{twice} the number of successful grasps compared to those from partial input alone. For dexterous grasping, we observe that the number of stable grasps sampled by DexGraspNet varies with object geometry. On average, we evaluate up to 100 random grasps per object and find that \coolname{} enables \textbf{twice} as many valid dexterous grasps compared to PartialDecomp. These improvements show that reconstructing scenes enables more reliable grasping and manipulation in cluttered environments.



\vspace{-9px}
\subsection{Real Robot Experiments}
\label{subsec:real_world}
We validate the real-world applicability of \coolname{} through experiments on a physical robot. Our experimental setup includes a Franka Emika Panda arm equipped with a wrist-mounted Intel RealSense D435 camera. 


\textbf{Dataset}: We evaluate our method on 15 distinct tabletop scenes collected in our lab, each containing 4 to 6 everyday objects, specifically chosen to minimize their likelihood of appearing in the training distribution of the methods.

\textbf{Baseline: }We compare against PartialDecomp and OctMAE in our evaluation. 

\textbf{Metrics}: For each scene, we capture an initial RGB-D image using the wrist-mounted camera and randomly select and execute kinematically feasible and collision-free grasps on each object in the scene. We measure the success rate as the percentage of objects that were successfully picked up and report the results in Table~\ref{tab:real-robot-metrics}.


\textbf{Results}: Our method achieves a success rate of \textbf{$\sim$73\%}, significantly outperforming OctMAE and grasps generated directly from the input partial point cloud alone. We show qualitative comparisons with OctMAE in Fig.~\ref{fig:lab_reconstruction_comparisons} and additional results in the supplementary. 
OctMAE struggles to produce plausible reconstructions on these everyday objects, often hallucinating geometry between distinct objects, leading to no valid grasps being generated for some objects. \coolname{} on the other hand, produces object reconstructions that align closely with the ground truth scene. We note that for \coolname{}, most grasping failures occur due to inaccuracies in the estimated object size, which subsequently leads to errors in registration. 
In general, \coolname{} allows robust manipulation of objects in cluttered real-world settings.

\begin{table}[htbp]
\centering
\small
\setlength{\tabcolsep}{6pt}
\renewcommand{\arraystretch}{1.2}
\begin{tabular}{lccc}
\hline
\textbf{} & Partial & OctMAE~\cite{iwase2024octmae} & \coolname{} \\
\hline
\textbf{Success Rate$\uparrow$} & $36.7 \pm 9.9$ & $59.6 \pm 15.3$ & $\textbf{73.3} \pm \textbf{15.2}$ \\ 
\hline
\end{tabular}
\caption{\textbf{Real robot} success rate (\%) measured as the percentage of objects picked successfully by the real robot for each method.}
\label{tab:real-robot-metrics}
\vspace{-26pt}
\end{table}

\section{Discussion}
\label{sec:discussion}


\paragraph{Limitations} Although our results are very promising, there are of course many failure modes in a composition of so many modules, which can have cascading effects on overall system performance.  We outline some failure modes and opportunities for improvement.
\textbf{Prompting and segmentation:} Occasionally the VLM fails to detect some of the object(s) in the image. Tuning the prompt mitigates the problem, as can prompting multiple ways for multiple hypotheses. 
\textbf{Segmentation: }Grounded-SAM occasionally segments parts of an object along with the full object which leads to multiple reconstruction hypotheses for the same object. We mitigate this partially using IoU-based de-duplication.
\textbf{Inpainting:} Our current inpainting strategy operates on a relatively isolated object, which removes some important context.  We mitigate this by slightly increasing the bounding box and adapting the model, but there is room for improvement.
\textbf{Image-to-3D:} Although remarkable, these models can sometimes fail to generate plausible reconstructions when given images from highly unusual viewpoints. 
\textbf{Scaling and registration:} Our scaling method is naive and would be improved in some cases by making it non-isotropic. Registration sometimes fails on uniformly-textured objects, where it is difficult to find distinctive features.  
\textbf{Seed value:} The performance of the image-to-3D and inpainting models may sometimes vary depending on the seed value selected. 



\paragraph{Conclusion \& Future Work}
We have presented a system that solves full-scene reconstruction from a single real-world RBG-D input in cluttered, occluded scenes with no assumptions about object categories. 
We have built on an incredibly strong foundation of existing general-purpose open-domain perception models and believe our approach will be able to adapt to and profit from future advances in such models.
One goal of this paper is to emphasize the importance of this problem for robot manipulation in real open-world environments and to encourage others to propose alternative solution strategies. We hope that overall advances in scene understanding in realistic manipulation settings will enable much more robust and capable robot manipulation systems.
One important strategy for making the system less error-prone is to move to a more generative setting with quantified uncertainty. 
If each module could generate multiple hypotheses, conditioned on its inputs, it would be possible to search for an interpretation that is collectively high-probability for all the modules.

\vspace{-10px}

\bibliographystyle{IEEEtran}
\bibliography{root}

\vfill

\end{document}